\documentclass{article}

\usepackage{amsthm}
\usepackage{amsmath}
\usepackage{amsfonts}
\usepackage{amssymb}
\usepackage{graphicx}
\usepackage{float}
\usepackage{verbatim}
\usepackage{enumerate}
\usepackage{mathtools}

\usepackage[acronym,hyperfirst=false]{glossaries}

\newtheorem{example}{Example}

\newtheorem{definition}{Definition}[section]
\newtheorem{remark}{Remark}

\newcommand{\Definition}[1]{Definition~\ref{#1}}
\newcommand{\Section}[1]{Section~\ref{#1}}

\newcommand{\Theorem}[1]{Theorem~\ref{#1}}
\newcommand{\Proposition}[1]{Proposition~\ref{#1}}
\newcommand{\Example}[1]{Example~\ref{#1}}

\renewcommand{\P}{\mathcal{P}}
\newcommand{\Not}{\textrm{\bf not\,}}
\newcommand{\head}{head}
\newcommand{\intersect}{\cap}
\newcommand{\union}{\cup}

\newcommand{\bodyp}{body^{+}}
\newcommand{\bodyn}{body^{-}}
\newcommand{\body}{body}

\newcommand{\KB}{\mathcal{K}}
\newcommand{\bfK}{\textit{\em \textbf{K}}}

\renewcommand{\P}{\mathcal{P}}
\newcommand{\KAK}{\textbf{KA}(\KB)}
\newcommand{\Katom}{\textbf{K}-atom}
\newcommand{\Katoms}{\textbf{K}-atoms}

\newcommand{\M}{\mathcal{M}}

\newcommand{\MM}{\langle M, M_1 \rangle}
\newcommand{\N}{\mathcal{N}}

\newcommand{\NN}{\langle N, N_1 \rangle}
\newcommand{\MN}{\langle M, N \rangle}
\newcommand{\ff}{{\bf f}}
\newcommand{\uu}{{\bf u}}
\renewcommand{\tt}{{\bf t}}
\newcommand{\mneval}[1]{(I, \MN, \MN)( #1  )}

\newcommand{\mmeval}[1]{\mneval{#1}}

\newcommand{\OO}{\mathcal{O}}
\newcommand{\OBO}[1]{{{\sf OB}_{\OO,#1}}}
\renewcommand{\P}{\mathcal{P}}

\renewcommand{\implies}{\supset}

\newcommand{\lfp}{{\bf lfp}}

\makeatletter
\newcommand\ineq{}
\DeclareRobustCommand{\ineq}{\mathrel{\mathpalette\rance@ineq\relax}}
\newcommand{\rance@ineq}[2]{%
  \vcenter{\hbox{%
    $\m@th#1\mkern1.5mu\underline{\mkern-1.5mu\bowtie\mkern-1.5mu}\mkern1.5mu$%
  }}%
}
\makeatother

\newcommand{\mknfmodels}{\models_{\mbox{\tiny{\sf MKNF}}}}
\usepackage[hidelinks]{hyperref}
\usepackage{orcidlink}

\usepackage[createShortEnv,conf={end,restate,no link to proof}]{proof-at-the-end}

\makeatletter
\renewcommand{\boxed}[1]{\text{\fboxsep=.2em\fbox{\m@th$\displaystyle#1$}}}
\makeatother

\newcommand{\argblank}{\underline{\hspace{.3cm}}}
\newcommand{\LL}{\mathcal{L}}

\renewcommand{\implies}{\Rightarrow}
\newcommand{\define}{\coloneqq}
\newcommand{\lcomp}[1]{{#1}^{\mathbf{\hspace{.075em}c}}}

\newcommand{\shortcite}[1]{\cite{#1}}

\usepackage{bm,color}

\title{Eliminating Unintended Stable Fixpoints for Hybrid Reasoning Systems}

\author{Spencer Killen\orcidlink{0000-0003-3930-5525} \and 
Jia{-}Huai You\orcidlink{0000-0001-9372-4371}\\Department of Computing Science, University of Alberta, Canada
\\\textbf{\{sjkillen, jyou\}@ualberta.ca}}

\begin{document}
\maketitle

\begin{abstract}
A wide variety of nonmonotonic semantics can be expressed as approximators defined under AFT (Approximation Fixpoint Theory). Using traditional AFT theory, it is not possible to define approximators that rely on information computed in previous iterations of stable revision. However, this information is rich for semantics that incorporate classical negation into nonmonotonic reasoning. In this work, we introduce a methodology resembling AFT that can utilize priorly computed upper bounds to more precisely capture semantics. We demonstrate our framework's applicability to hybrid MKNF (minimal knowledge and negation as failure) knowledge bases by extending the state-of-the-art
approximator.
\end{abstract}


\section{Introduction}\label{section-introduction}
Stable revision is the core of approximation fixpoint theory (AFT)~\cite{denecker2000approximations,DeneckerMT04}.
The {\em stable (revision) operator} is defined in terms of an underlying operator (the approximator) and
when applied in nonmonotonic reasoning, this approximator maps three-valued logical interpretations\footnote{The three values being false, undefined and true\\Proofs have been attached in Appendix \ref{appendix}}
~to three-valued interpretations with fewer undefined atoms.
Stable revision enhances approximators with the capacity to rule out cyclically justified inferences.
The fixpoints of stable revision operators, called {\em stable fixpoints}, can characterize the stable and partial stable semantics (e.g., the well-founded semantics) of many nonmonotonic reasoning systems~(e.g.\ \cite{aftagg,aftjust}). 

When applied to logic programming, the stable revision operator begins with the interpretation that assigns all atoms to be undefined, and gradually assigns atoms a truth value (either true or false).
To detect cyclic justifications, the stable revision operator computes possibly true atoms iteratively.
In the middle of computation, it's not possible to discern between atoms that will be computed as possibly true on a subsequent iteration and atoms that were established as false on a prior iteration of stable revision.
As a result, the underlying approximator cannot make inferences that safely rely on the falsity of atoms.
This limits inference power with nonmonotonic logics that incorporate classical negation as classical negation requires proof of falsity.
When formulated using AFT,  stable operators cannot reason about the
complement of an approximation's upper bound. This ``negative information'' is
useful when reasoning with semantics that mix nonmonotonic and classical
reasoning. 
To be a bit more concrete, (the full details will be given later in the paper) given a pair $(T,P)$ on a bilattice under precision ordering and an approximator $o$, the stable revision operator $S$ is defined by a pair of least fixpoints of operators over the underlying lattice:
        $$S(o)(T, P) = (\lfp (o{(\cdot, P)}_1), \lfp (o{(T,\cdot)}_2)$$
        where $\lfp (o{(\cdot, P)}_1)$, with $P$ fixed, iteratively computes a new lower bound, which is projected as the first element of the resulting pair generated by operator $o$; similarly for $\lfp (o{(T,\cdot)}_2)$. In the context of nonmonotonic reasoning systems, a pair $(T,P)$, which we call an {\em approximation}, represents a partial interpretation, where $T$ is the set of atoms that are assigned to true and $P$ is the set of atoms that are possibly true; hence $P$ serves as an upper bound whose complement is the set of false atoms. Notice, however, since the new upper bound is computed ground up, it is erroneous to assume that the complement of such an intermediate set represents false atoms. In other words, the ``established'' false information w.r.t.\ the given $P$ is not accessible in the computation of a new upper bound.
There is a possibility for this information to be provided via an extra parameter.

The primary contribution of this work is a formulation of stable revision where
approximators have an additional parameter that encodes information computed in a previous iteration of stable revision (before all atoms are set to false).
We use this additional parameter to make more powerful inferences and to ultimately remove undesired fixpoints from the stable revision operator.

The theory of fixpoint operators we employ is not a strict realization of
Denecker et al.'s consistent and symmetric AFT \shortcite{denecker2000approximations}, but rather it mirrors the work of Liu and
You~\shortcite{liuyou2021}.
Liu and You recognize that any approximator, whose stable fixpoints are
precisely \mbox{3-valued} hybrid MKNF models (minimal knowledge and negation as failure), is super-polynomial\footnote{determining whether a knowledge base has an MKNF model is NP-hard \cite{liuthreevalued2017}}.
For this reason, Liu and You use stable revision to capture a superset of their intended
models (the three-valued MKNF models) and they provide a polynomially-checkable property that can be used
to filter unintended stable fixpoints.
It is desirable to have fewer unintended stable fixpoints as they sometimes
block stable revision from computing the well-founded model. 

Also, our proposed framework can express operators that were out of reach of approximation fixpoint theory. In some previous work,  e.g., in formulating constraint propagation for DPLL solvers \cite{Ji17,killenunfounded}, well-founded operators for hybrid MKNF knowledge bases already make inferences in terms of the ``previous state'' of the approximation. In this paper, 
we formulate a general method for stable revision to make such inferences. In fact, this generalization leads to stable revision that is more powerful (in the sense of generating more true/false atoms for the intended applications)
than any of the previous operators.

\begin{example}\label{example-running}
    Suppose we have a hybrid MKNF knowledge base comprised of a first-order
    theory $\mathcal{O}$ that simply asserts
    that the atom $c$ is false and the following set of nonmonotonic rules, where positive atoms are represented by modal $\bfK$-atoms. 
    \begin{align*}
         & 1.\hspace{1.5em} \bfK a\hspace{0.25em} \leftarrow \Not a'.
         & 2.\hspace{1.5em} \bfK a' \leftarrow \Not a.
        \\
         & 3.\hspace{1.5em} \bfK c\hspace{0.25em} \leftarrow \bfK a,~ \Not b.
    \end{align*}
     According to 3-valued MKNF
    \cite{knorrlocal2011}, this knowledge base has a single \mbox{3-valued} model (which happens to be 2-valued): the model that assigns
    $a'$ to true and every other atom to false.\footnote{For simplicity, we sometimes drop the $\bfK$ symbol in reference to an atom, but we keep it in rules}.
    Rules 1 and 2 assert that either $a$ or
    $a'$ is true.
    Rule 3 is of particular interest: Due to the first-order theory, the atom
    $c$ is false, therefore the rule's body must also be false, i.e.\ either
    $a$
    must be false or $b$ must be true.
    A pair $(T,P)$, which we call an {\em approximation}, represents a partial interpretation, where $T$ is the set of atoms that are assigned to true and $P$ is the set of atoms that are possibly true; hence $P$ serves as an upper bound whose complement is the set of false atoms.
To compute the least stable fixpoint, we start with the least element $(\emptyset, \Sigma)$ of the underlying bilattice, where $\Sigma$ denotes the set of all atoms.
If we adopt the approximator based on 
Knorr et al.'s alternating fixpoint construction \shortcite{knorrlocal2011}, Liu and You's richer approximator $\Psi$ \cite{liuyou2021} (cf. Def.\,8 of the paper), or an approximator based on the well-founded operators~\cite{Ji17,killenunfounded}, then the computed least stable fixpoint is the pair $(\emptyset, \{a, a'\})$, which does not correspond to a model because rule 3 is not satisfied. 

Assume we have access to the false atoms established in the previous iteration of stable revision, in this case, $\{ b, c \}$. Then the atom $a$ can be inferred as false in subsequent iterations since $c$ and $b$ were established as false in a prior iteration. This leads to the eventual least stable fixpoint, $(\{a'\}, \{a'\})$, which corresponds to the well-founded model. 

One may attempt to construct the set of false atoms in the same iteration of 
computing a new upper bound by complementing the upper bound. This may generate erroneous results. Consider, for example,  adding the following rule to the above rule set:
    $$4.\hspace{1.5em} \bfK b\hspace{0.25em} \leftarrow \bfK a.$$
    Now, instead of having a well-founded model, the knowledge base has two \mbox{3-valued} models: the same one as before and a model that assigns $a$ and $b$ to true. 
    Because the knowledge base does not have a well-founded model, we would like the computed least stable fixpoint to be less defined than both of the models.\footnote{Such a fixpoint is useful in some applications, e.g., in grounding rules.}
    Now, let $o'$ be an approximator. 
    Then, 
in the first step of computing 
$\lfp (o'{(\emptyset,\cdot)}_2)$, namely in invoking $o'(\emptyset,\emptyset)$,
because $b$ is false (as $b \not \in \emptyset$) and $c$ is false, we infer that $a$ is false. Thus, we computed the least stable fixpoint $(\{a'\}, \{a'\})$ and erroneously claim that it is the well-founded model.
\end{example}

In the example above, we demonstrate how leveraging false information can increase the
inference power of stable revision.
Stable revision is comprised of two levels of iteration: on the outer layer,
the set of possibly-true atoms shrinks with each iteration.
However, this outer layer is formed from an iterative fixpoint process that
computes the set of possibly-true atoms in the opposite direction
starting with the empty set.
From the perspective of this ``inner'' operator, it is impossible to discern whether an
atom is false or whether an atom has yet to be computed as \mbox{possibly-true}.
We can circumvent this limitation by keeping track of the set of possibly-true
atoms that were computed in the last iteration of stable revision.
Because the set of possibly-true atoms shrinks each iteration, the inner
fixpoint computation will compute at most the atoms that were computed last
iteration.
Therefore the atoms in the complement of the previously computed set $P$ can safely be treated as
false.

We formulate a framework for defining approximators capable of correctly determining whether an atom is false during the computation of the set possibly-true atoms \--something that traditional approximators cannot do.
Surprisingly, our extension does not require new theory. We simply modify the underlying bilattice on which the approximators operate.
There is opportunity for approximators described using this framework to have fewer unintended stable fixpoints if such fixpoints exist.
We demonstrate the utility of this framework by lifting our well-founded operator for hybrid MKNF knowledge bases~\cite{workshop} to be an approximator --this was not possible before.

We organize the paper as follows. 
As preliminaries, \Section{section-prelim}
details lattice theory and the notation adopted throughout this work and 
\Section{section-prelim-approx} covers approximators and stable revision as used in this work.
In \Section{section-main-theory}, we introduce the \textit{recurrent
    approximators}, approximators defined over a \textit{tetralattice}, a
bilattice
formed from a bilattice. These operators operate on 4-tuples (pairs of pairs) and provide a theoretical backing on AFT with an extra parameter of pairs. Then, we turn our attention to an application. 
\Section{section-prelim-hmknf} gives an overview of the logic of MKNF and
hybrid MKNF knowledge bases, and 
in \Section{section-main}, we demonstrate the utility of this family of approximators by lifting and
increasing the precision of the AFT operator defined by Liu and
You~\cite{liuyou2021} for hybrid MKNF knowledge bases.
This approximator widens the set of knowledge bases which have a known polynomial algorithm to compute their well-founded models.

\section{Preliminaries}\label{section-prelim}

We recite common theory of lattices~\cite{romanlattices2008} to establish the
notation used throughout this work.
A \textit{poset} ${\langle S, \preceq_{\alpha} \rangle}$ is a relation
$\preceq_{\alpha}$ over a set of elements $S$ that satisfies:
\textit{reflexivity}  ($\forall x \in S, x \preceq_{\alpha} x$),
\textit{transitivity} ($\forall x, y, z \in S, \textit{having both } x
\preceq_{\alpha} y$ and $y \preceq_{\alpha} z$ implies $x \preceq_{\alpha} z$),
and
\textit{antisymmetry} ($\forall x, y \in S, \textrm{ if } x \preceq_{\alpha} y$
and $y \preceq_{\alpha} x$ then $x = y$).
We refer to a poset ${\langle S, \preceq_{\alpha} \rangle}$ simply by $S$ when
$\preceq_{\alpha}$ is clear from context.
Given a poset $S$, we call an element $x \in S$ an \textit{upper bound} (resp.
a
\textit{lower bound}) of a subset $Q \subseteq S$ if $\forall y \in Q, y
    \preceq x$ (resp. $\forall y \in Q, x \preceq y$).
An upper bound of $Q$ w.r.t. a poset $\langle S, \preceq_{\alpha} \rangle$ is a
\textit{least upper bound}, denoted $lub(Q)$ (resp. \textit{greatest lower
    bound}, denoted as $glb(Q)$) if it is a lower bound of the set of all upper
bounds of $Q$ (resp. an upper bound of the set of all lower bounds of $Q$).
A poset $\langle {\LL}, \preceq_{\alpha} \rangle$ is a \textit{complete lattice}
if
every subset $S \subseteq {\LL}$ has a least upper bound and a greatest lower
bound.
For a complete lattice $\langle \mathcal{{\LL}}, \preceq_{\alpha} \rangle$ we
denote $glb(\mathcal{{\LL}})$ as $\bot_{\preceq_{\alpha}}$ and
$lub(\mathcal{{\LL}})$
as $\top_{\preceq_{\alpha}}$ when ${\LL}$ is clear from context or simply as
$\bot$
and $\top$ when the relation is unambiguous.

An \textit{operator} over a complete lattice $\langle {\LL}, \preceq_{\alpha}
    \rangle$ is a function $o(x): {\LL} \rightarrow {\LL}$.
The operator is $\preceq_{\alpha}$-monotone (resp. $\preceq_{\alpha}$-antitone)
if $\forall x, y \in {\LL}$ whenever $x
    \preceq_{\alpha} y$ we also have $o(x) \preceq_{\alpha} o(y)$ (resp. $o(y)
    \preceq_{\alpha} o(x)$).
An operator is $\preceq_{\alpha}$-monotone increasing (resp. decreasing) if
$\forall x, x \preceq_{\alpha} o(x)$ (resp. $\forall x, o(x) \preceq_{\alpha}
    x$).

An element of a complete lattice $x \in {\LL}$ is a fixpoint of an operator $o$
if
$o(x) = x$.
The set of all fixpoints of a $\preceq_{\LL}$-monotone operator $o$ on a lattice
$\langle {\LL},
    {\preceq_{{\LL}}} \rangle$ forms a complete
lattice~\cite{tarskilatticetheoretical1955}.
We call the greatest lower bound of this lattice \textit{the least fixpoint}
and denote it as $\lfp_{\preceq_{{\LL}}}{o}$.
This element can be constructed by iteratively applying $o$ to
$\bot_{\preceq_{{\LL}}}$.
We denote the cartesian product of two sets $S$ and $D$ with $S \times D$ or
$S^2$ if $S = D$, that is,
\begin{align*}
    S \times D & \define \{ (s, d) ~|~ s \in S, d \in D \} \\
    S^2        & \define \{ (s, a) ~|~ s \in S, a \in S \}
\end{align*}

Given a lattice $\langle {\LL}, \preceq_{{\LL}} \rangle$, its induced bilattice
\cite{denecker2000approximations} consists of the two complete lattices
$\langle {\LL}^2, \preceq^2_p
    \rangle$ and $\langle {\LL}^2, \preceq^2_t \rangle$. These are the lattices
formed
from the two orderings $\preceq^2_p$ and $\preceq^2_t$ such that for each $x,
    y, z, w \in {\LL}$
\begin{itemize}
    \item $(x, y) \preceq^2_p (z, w)$ iff $x \preceq_{{\LL}} z$ and $w
              \preceq_{{\LL}}
              y$  (the precision-ordering)
    \item $(x, y) \preceq^2_t (z, w)$ iff $x \preceq_{{\LL}} z$ and $y
              \preceq_{{\LL}}
              w$  (the truth-ordering)
\end{itemize}

We denote the powerset of a set $S$, as $\wp(S)$.
We use subscript notation to denote the projection of particular components of
a
tuple, for example, given an operator $o(T, P): {\LL}^2 \rightarrow {\LL}^2$, we
have
$o(T, P) = (o(T, P)_1, o(T, P)_2)$ and $o(T, P)_{2, 1} = (o(T, P)_2, o(T,
    P)_1)$.

We create partial functions by using a ``$\cdot$'' in place of arguments to be
filled in, that is, for an operator $o(T, P): {\LL}^2 \rightarrow {\LL}^2$, we
write
$o(\cdot, P)$ (resp. $o(T, \cdot)$) to mean $\lambda x.~ o(x, P)$ (resp.
$\lambda x.~ o(T, x)$).
Naturally, if a ``$\cdot$'' is used within a function application that is then
projected, the projections are included within the body of the lambda
abstraction, for example,
\begin{align*}
    f(x, \cdot)_1 & = \lambda y.~ (f(x, y)_1) ~~~ &  & (\textit{where $f(x, y):
        {\LL}^2
        \rightarrow {\LL}^2$})
\end{align*}
This makes it possible to write $\lfp\preceq_{\LL} o(T, \cdot)_1$.

We use an ``$\argblank$'' in a function's signature to signify that an argument
is
consumed, but not used in the body of the function
, for example,
\begin{align*}
    f(x, \argblank): {\LL}^2 \rightarrow {\LL}^2 & = \lambda (x, \argblank).~ (x,
    x)
    \\
    f(1,2)                                     & = (1, 1)
\end{align*}

For convenience and ergonomics, we may write 4-tuples as a pair of 2-tuples or
as a tuple with four members.
As a general rule, we consider two tuples to be equivalent if they are equal
when all nested tuples are ``flattened''. For example, the following
equivalences hold.
\begin{align*}
    {\LL}^4            & = {\LL}^2 \times {\LL}^2	= {({\LL}^2)}^2 \\
    (T, F, U, P)      & = ((T, F), (U, P))                   \\
    f((T, F), (U, P)) & = f(T, F, U, P)
\end{align*}

\section{Approximators}\label{section-prelim-approx}

This work adopts the generalized framework AFT described by Liu and You~\shortcite{liuyou2021}.
Here, our primary focus is stable revision and we do not require that approximators are symmetric or consistent.
Stable revision captures a superset of our intended models: that is, a stable fixpoint may not be an intended model, but all
intended models are stable fixpoints. This difference from Denecker et al.'s AFT~\shortcite{denecker2000approximations} does not prevent us from
applying the framework to characterize intended semantics. Instead, stable
revision is coupled with a property to check whether a stable fixpoint corresponds to an intended model.
We refer to stable fixpoints that do not satisfy this property as unintended stable fixpoints.
It is desirable to have fewer unintended stable fixpoints and approximators are improved if they're tuned to have fewer.

Our motivation for not adhering to consistent AFT~\cite{denecker2000approximations} is multifaceted.
Semantics that tightly couple classical and nonmonotonic reasoning require
special treatment of inconsistency~\cite{RRaft}. Non-symmetric AFT is better
suited to deal with the inconsistencies that naturally arise from classical
reasoning.
Denecker et al.\ \shortcite{denecker2000approximations} initially suggested that AFT
could be formulated without the symmetry requirement imposed on approximators,
however, later developments of the framework make heavy use of these
restrictions~\cite{DeneckerMT04}.
Simplicity is another motivating factor of ours \-- our use of stable revision is
surprisingly simple since it only relies on operator monotonicity and we thus
can work with the complete bilattice rather than a consistent chain-complete
subset of the bilattice.

It is often the case that the notion of an operator targeted by approximations
is dropped and AFT is used.
The primary focus of this work is a formulation of stable revision that
leverage previously computed upper bounds to obtain more precise fixpoints.
We formulate and utilize mechanisms inspired by
AFT~\cite{denecker2000approximations}, however, we deviate in a few key ways.
We do not enforce symmetry with approximators or require that approximators map
to consistent approximations. We use stable revision alone as a means of
characterizing semantics.

While the theory in this work stands on its own, it is also a preliminary step
towards a generalized AFT framework that does not rely upon consistency or
symmetric operators.
Namely a step that gauges the applicability of such a framework.

If an approximator is not consistent, then its stable operator may
have fewer fixpoints than the approximator~\cite{RRaft}.
For this reason, we limit our concern to fixpoints computed by stable revision
and disregard the fixpoints of approximators.
A fixpoint of an approximator may only correspond to an intended model if it is
also a fixpoint of stable revision.
Because these stable fixpoints capture a superset of our intended models,
removing stable fixpoints that are unintended models is of particular interest.
The approximator defined by Liu and You~\cite{liuyou2021}, makes use of
negative information generated by classical theories to perform unit
propagation on nonmonotonic rules.
However, this extension is limited to the negative knowledge that is
immediately derivable from classical theories.

This type of propagation has proven to be difficult to do with existing AFT
methods.
In essence, each iteration of stable revision ``resets'' the set of ``possibly
true`` atoms by assigning them all to be false.
Because some of these atoms may become possibly-true again in a successive
iteration, we do not have access to atoms that are well-established to be
false, this information is discarded by stable revision.
When computing from the least fixed point, the set of atoms the were not
computed to be possibly-true is precisely the set of well-established false
atoms.

In this work, we extend AFT to enable the construction of operators with access
to information computed during prior iterations of stable revision so that we
can access this false information.
Our extension keeps within the traditional AFT (with the exception that we
allow for approximators that are not symmetric), and thus requires only a few
new definitions.


We introduce the definitions of approximators and stable revision~\cite{liuyou2021}.
\begin{definition}\label{defn-deterministic}
    An {\em approximator} is a $\preceq_{p}^2$-monotone operator ${o(T, P):
        {\LL}^2
        \rightarrow {\LL}^2}$
    on the complete lattice $\langle {\LL}^2, \preceq_{p}^2 \rangle$
\end{definition}

\begin{definition}
    Given an approximator $o(T, P)$, the \em{stable revision operator} $S(o)$
    is defined as follows:
    \begin{align*}
        S&: \Big( {\LL}^2 \rightarrow {\LL}^2 \Big) \rightarrow
        {\LL}^2 \rightarrow
        {\LL}^2 \\
        S(o)(T, P) &\define (\lfp_{\preceq_{{\LL}}} (o{(\cdot, P)}_1), \lfp_{\preceq_{{\LL}}}
            (o{(T,
        \cdot)}_2))  
    \end{align*}
\end{definition}
For an approximator $o$, we refer to fixpoints of $S(o)$ as \mbox{\em stable fixpoints}.  Since the operator $o$ is ${\preceq^2_p}$-monotone, it is easy to check that 
both operators $o(\cdot, P)_1$ and $o(T, \cdot)_2$ are ${\preceq_{\LL}}$-monotone, so stable revision is well-defined.

\section{Recurrent Approximators}\label{section-main-theory}

The framework described in~\Section{section-prelim-approx} cannot construct some well-founded operators defined for hybrid MKNF knowledge bases \cite{Ji17,killenunfounded}.
These well-founded operators are formulated as families of approximators.
Each approximator in this family is induced by an approximation $(T, P)$ and the approximator always computes an approximation $(T^*, P^*)$ that is more precise than $(T, P)$, that is, $(T, P) \preceq^2_p (T^*, P^*)$. The less precise approximation $(T, P)$ contains ``stale'' information that was computed with the same operator on an earlier iteration.
This property enables the well-founded operators to safely reason about the falsity of atoms \-- which is difficult to do with approximators.

We describe our process to get approximators (\Definition{defn-deterministic}) to embed stale approximations by modifying the underlying lattice so that elements store an older approximation in addition to the current one.
This modification has no impact on approximators that do not utilize this information and allows us to define approximators with fewer unintended stable fixpoints.

Throughout the remainder of this work, we assume that every complete lattice $\langle \LL,
    \preceq_{\LL} \rangle$ has a complement operation, denoted as $\lcomp{a}$ and that satisfies the
following two properties.
\begin{align*}
    i.~ \forall a \in \LL,~ \lcomp{(\lcomp{a})} = a
    &&ii.~ \forall a, b \in \LL,~ a \preceq_{\LL} b \iff \lcomp{b} \preceq_{\LL}
              \lcomp{a}
\end{align*}
In a powerset lattice, a natural choice for this operation is the set complement operation.
Some lattices have many or no possible complement operations, thus it appears limiting, however, this operation is not necessary to apply our
theory. We rely upon it only for simplicity. One can instead define orderings differently so that the criteria of the complement is satisfied.
In Appendix~\ref{appendix-no-complement}, we briefly describe how this condition may be dropped.

We intend to isolate a family of approximators defined on a ``bilattice formed from a bilattice'' that can be used to propagate information
from previous iterations.
First, we formally describe this lattice.
\begin{definition}\label{orderings}
    Given a complete lattice $\langle \LL, \preceq_{\LL} \rangle$ we construct its
    bilattice ${\langle {\LL}^2, \preceq^2_t \rangle}$, $\langle {\LL}^2, \preceq^2_p \rangle$ then define the
    following pair of complete lattices which we refer to collectively and individually as a tetralattice.
    \begin{align*}
        \langle {\LL}^4, \preceq^4_t \rangle, \langle {\LL}^4, \preceq^4_p
        \rangle
    \end{align*}
    A tetralattice is the bilattice formed from turning ${\langle{\LL}^2,  \preceq^2_t\rangle}$ into a bilattice.
    The orderings $\preceq^4_t$ and $\preceq^4_p$ are naturally defined, but we breakdown their definition below for convenience.
    For the ordering $\preceq_t^4$ and any two 4-tuples $${(T, F, U, P), (T', F', U', P') \in {\LL}^4}$$the following three expressions are equivalent\,\footnote{A fourth equivalence could be added that is defined in terms of $\preceq_p^2$, however, we do not make use of such a formulation.}
    \begin{itemize}
        \item $((T, F), (U, P)) \preceq_t^{4} ((T', F'), (U', P'))$,
        \item $(T, F) \preceq^2_{t} (T', F') \land (U, P) \preceq^2_{t} (U',
                  P')$,
              and
        \item $T \preceq_{\LL} T' \land F \preceq_{\LL} F'	~\land U \preceq_{\LL} U'
                  \land P
                  \preceq_{\LL} P'$
    \end{itemize}
    For $\preceq^4_p$, the following are equivalent
    \begin{itemize}
        \item $((T, F), (U, P)) \preceq_p^{4} ((T', F'), (U', P'))$,
        \item $(T, F) \preceq^2_{t} (T', F') \land (U', P') \preceq^2_{t} (U,
                  P)$,
        \item $(T, P) \preceq^2_{p} (T', P') \land (F, U) \preceq^2_{p} (F',
                  U')$,
              and
        \item $T \preceq_{\LL} T' \land F \preceq_{\LL} F' \land U' \preceq_{\LL} U \land
                  P'
                  \preceq_{\LL} P$
    \end{itemize}
\end{definition}
Intuitively, we take the process applied to $\langle L, \preceq_{\LL} \rangle$
to obtain $\langle {\LL}^2, \preceq^2_t
    \rangle$ and $\langle {\LL}^2, \preceq^2_p \rangle$, then we apply it to the
lattice $\langle {\LL}^2, \preceq^2_t
    \rangle$.
The result is a pair of complete lattices because bilattices are complete lattices~\cite{fittingfixpointsurvey}.
\begin{definition}
    A \textit{recurrent operator} $o(T, F, U, P): {\LL}^4 \rightarrow {\LL}^4$ is an
    operator on the bilattice ${\LL}^4$ such that
    \begin{align*}
        o{(T, F, U, P)}_{2,3} = \big( \lcomp{P}, \lcomp{T} \big)
    \end{align*}
    A $\preceq^{4}_{p}$-monotone recurrent operator is called a \textit{\mbox{recurrent approximator}}.
\end{definition}

Note that $o{(T, F, U, P)}_{2,3}$ is of type $L^4 \rightarrow L^2$.
Intuitively, we are fixing $o{(\cdot, \cdot, \cdot, \cdot)}_{2}$
and $o{(\cdot, \cdot, \cdot, \cdot)}_{3}$ (sometimes referred to as ``the inner components'') to be the functions ${\lambda
            (\argblank, \argblank, \argblank, P):~ \lcomp{P}}$ and
$\lambda (T,
            \argblank, \argblank, \argblank):~ \lcomp{T}$ respectively.
To construct a recurrent operator, we only need to define $o{(\cdot, \cdot,
            \cdot, \cdot)}_{1, 4}$, a traditional approximator that
additionally receives
an older computation of $T$ and $P$ (in complement form) and returns a new approximation $(T', P')$.
The utility of these previous states is not fully apparent until embedded in the
stable revision operator.

The least element of the lattice $\langle {\LL}^4, \preceq^4_p \rangle$ is the
pair $((\bot_{\LL}, \bot_{\LL}), (\top_{\LL}, \top_{\LL}))$ which is equivalant to
$(\bot_{\preceq^2_{t}}, \top_{\preceq^2_{t}})$.

In essence, the operator $o(T, F, U, P)_{1, 4}$ is an approximator
defined for the bilattice ${\LL}^2$ that has been ``lifted'' to ${\LL}^4$ whereas
the operator $o(T, F, U, P)_{2, 3}$ functions as the recurrent portion of the
operator which will store old information during stable revision.
For this reason, we need only concern ourselves with the definition of $o(T, F,
    U, P)_{1, 4}: {\LL}^4 \rightarrow {\LL}^2$.

In the following lemma, we show that the conditions for
$\preceq^{4}_{p}$-monotonicity for a recurrent operator can be relaxed slightly.
\begin{lemmaE}[]\label{relaxed-monotonicity}
    For a tetralattice ${\langle \LL^4, \preceq^4_{p} \rangle}$, a recurrent
    operator $o(T, F, U, P)$ is $\preceq^{4}_{p}$-monotone iff
    for each ${(T, F, U, P), (T', F', U', P') \in L^4}$ s.t. 
    $$(T, F, U, P) \preceq^4_{p} (T', F', U', P')$$ We have ${o{(T, F, U, P)}_{1,4} \preceq^2_p  o{(T', F', U', P')}_{1,4}}$
\end{lemmaE}
\begin{proofE}
    ($\Rightarrow$) trivial.

    ($\Leftarrow$) It's sufficient to show $o{(T, F, U, P)}_{2, 3} \preceq^2_p
        o{(T', F', U', P')}_{2,3}$.
    We have $o{(T, F, U, P)}_{2} = \lcomp{P}$ and $o{(T', F', U', P')}_{2} = \lcomp{P'}$.
    Clearly since $P' \preceq_{\LL} P$, we have $\lcomp{P} \preceq_{\LL} \lcomp{P'}$.
    We have $o{(T, F, U, P)}_{3} = \lcomp{T}$ and $o{(T', F', U', P')}_{3}
        = \lcomp{T'}$.
    From $T \preceq_{\LL} T'$, we have $\lcomp{T'} \preceq_{\LL} \lcomp{T}$.
    We conclude  $(\lcomp{P}, \lcomp{T}) \preceq^2_p (\lcomp{P'}, \lcomp{T'})$.
\end{proofE}

For convenience, we repeat the definition of the stable revision operator using
$\langle L^2, \preceq_t^2 \rangle$ as the underlying lattice in place of
$\langle L, \preceq_{\LL} \rangle$.
\begin{align*}
    S&: \Big( {\LL}^4 \rightarrow {\LL}^4 \Big) \rightarrow
    {\LL}^4 \rightarrow
    {\LL}^4\\
    S(o)(T, F, U, P) &\define (\lfp_{\preceq^2_{{t}}} (o{(\cdot, (U, P))}_{1, 2}),
    \lfp_{\preceq^2_{{t}}} (o{((T, F),
            \cdot)}_{3,4}))
\end{align*}

\begin{propositionE}[]\label{S-monotone}
    For a complete lattice $\langle \LL^4, \preceq^4_{p} \rangle$ and a
    $\preceq^4_p$-monotone operator $o(T, F, U, P)$, the stable revision
    operator
    $S(o)$ is a recurrent approximator.
\end{propositionE}
\begin{proofE}
    We show (i) that $S(o)$ is a recurrent operator and then (ii) that $S(o)$
    is $\preceq^4_p$-monotone.
    (i) Let $(T, F, U, P) \in \LL^4$.
    The functions $o{(\cdot, (U, P))}_{2}$ and $o{((T, F), \cdot)}_{3}$ are
    constant, therefore
    \begin{align*}
        S(o)(T, F, U, P)_{2,3} = (\LL\setminus P, \LL \setminus T)
    \end{align*}

    (ii) By \cite{tarskilatticetheoretical1955} and
    \cite{denecker2000approximations} the $S(o)$ operator is well-defined, that
    is,
    $o$ has fixpoints that exist when $o$ is monotone.
    Let $(T, F, U, P), (T', F', U', P') \in L^4$ such that $(T, F, U, P)
        \preceq^4_p (T', F', U', P')$.
    It is sufficient to show
    \begin{enumerate}[(a)]
        \item $\lfp_{\preceq^2_{{t}}} (o{(\cdot, (U, P))}_{1, 2})  \preceq^2_t
            \lfp_{\preceq^2_{{t}}} (o{(\cdot, (U', P'))}_{1, 2})$
        \item $\lfp_{\preceq^2_{{t}}} (o{((T', F'), \cdot)}_{3,4})  \preceq^2_t
            \lfp_{\preceq^2_{{t}}} (o{((T, F), \cdot)}_{3,4})$
    \end{enumerate}
    (a) Let $x = \lfp_{\preceq^2_{{t}}} (o{(\cdot, (U', P'))}_{1, 2})$. By the $\preceq^4_p$-monotonicity of $o$, we have
    \begin{align*}
        o(x, (U, P)) \preceq^2_t o(x, (U', P'))
    \end{align*}
    Here, $x$ is a prefixpoint of $o(\cdot, (U, P))$. $\lfp_{\preceq^2_{{t}}} (o{(\cdot, (U, P))}_{1, 2})$ corresponds to the least prefixpoint of $o$ \cite{tarskilatticetheoretical1955}, thus
    $\lfp_{\preceq^2_{{t}}} (o{(\cdot, (U, P))}_{1, 2}) \preceq^2_t x$.
    A nearly identical procedure can be used to show that  $\lfp_{\preceq^2_{{t}}} (o{((T', F'), \cdot)}_{3,4}) \preceq^2_t
    \lfp_{\preceq^2_{{t}}} (o{((T, F), \cdot)}_{3,4})$.
    We conclude that $S(o)$ is $\preceq^4_p$ monotone and with (i) it is a recurrent approximator.
\end{proofE}

It is convenient to have notation to map between tuples in $\LL^2$ and $\LL^4$.
We define the following mappings.
\begin{align*}
    {(T, F, U, P)}_{1, 4}  = (T, P)                       
    &~~~~~{(T, P)}^4             = (T, \lcomp{P}, \lcomp{T}, P)
\end{align*}

We give a simplistic example to demonstrate the mechanics of our definitions.
\begin{example}
    Let $\langle \LL, \preceq_{\LL} \rangle$ be a complete lattice where $\LL =
        \{\bot,~ +,~ \top\}$, $\preceq_S$ is a linear order where $\bot
        \preceq_{\LL} +
        \preceq_{\LL} \top$, and we use complement operation $\lcomp{\bot} = \top$, $\lcomp{+} = +$.
    $\alpha \in
    {\LL}$.
    \begin{align*}
        \lcomp{\alpha} \define \left\{\begin{array}{ll}
                                          \top & \textrm{iff $\alpha = \bot$}
                                          \\
                                          +    & \textrm{iff $\alpha = +$}
                                          \\
                                          \bot & \textrm{iff $\alpha = \top$}
                                      \end{array}\right.
    \end{align*}
    First, we define a traditional approximator $o$ over the bilattice ${\langle {\LL}^2,
                \preceq^2_p \rangle}$ to be the identity function.
    Clearly, this operator is $\preceq^2_p$-monotone.
    We can easily lift this operator to the tetralattice by defining $o(T, F, U, P)_{2, 3}$.
    By \Proposition{S-monotone}, this operator is $\preceq^4_p$-monotone, therefore it is a recurrent approximator.

    Here the least stable fixpoint, $\lfp_{\preceq^2_p} S(o)$ is $(\bot, +)$.
    Suppose we deem every stable fixpoint $(T, P)$ such that $P = +$ as an ``unintended model''.
    We wish to remove these stable fixpoints in favour of stable fixpoints that
    are more precise w.r.t. $\preceq^2_p$.
    However, we wish to keep every other fixpoint of $S(o)$.
    While this simple example is possible with an approximator over $L^2$, we demonstrate how an approximator defined over $L^4$ can achieve this using its additional parameters.
    We define a recurrent approximator over $\langle {\LL}^4,
        \preceq^4_p \rangle$
        \begin{align*}
        o(T, F, U, P)_{1, 2, 3} & \define (T, \lcomp{P},~ \lcomp{T})
        \\
        o(T, F, U, P)_{4}    & \define \left\{ \begin{array}{ll}
                                                   \bot & \textrm{if $\lcomp{F}
                                                           = +$}
                                                   \\
                                                   P    & \textrm{otherwise}
                                               \end{array}\right.
    \end{align*}
    \begin{align*}
        o(T, F, U, P)_{1, 2, 3}  \define (T, \lcomp{P},~ \lcomp{T}) \hspace{0.5cm}
        o(T, F, U, P)_{4}    & \define \left. \begin{array}{ll}
                                                   \bot  \textrm{ (if $\lcomp{F}
                                                           = +$)}
                                                        \textrm{ otherwise $P$ }
                                               \end{array}\right.
    \end{align*}
    Note that the least and greatest stable fixpoints are the same as if $o$ were the identity function
    \begin{align*}
        S(o)(\bot, \bot, \top, \top) = (\bot, \bot, \top, \top) \hspace{1cm}
        S(o)(\top, \top, \bot, \bot) = (\top, \top, \bot, \bot)
    \end{align*}
    However, for any $(T, F, U, +) \in \LL^4$ we have that $S(o)(T, F, U, +) \preceq^4_p (T, F, U, +)$.
\end{example}

In the coming sections, we define an approximator for hybrid MKNF
knowledge bases so that we can give a more concrete application of this framework.
For these sections, we narrow in on one particular type of tetralattice.
\begin{definition}
    A \textit{powerset tetralattice} $\langle \wp(\LL)^4, \preceq_p^4 \rangle$ is the tetralattice formed from a powerset
    lattice ${\langle \wp(\LL), \subseteq \rangle}$ using ${\lcomp{\alpha} = \LL
    \setminus \alpha}$ as the complement operation.
\end{definition}

\section{Hybrid MKNF Knowledge Bases}\label{section-prelim-hmknf}
MKNF is a modal autoepistemic logic defined by Lifschitz
\shortcite{lifschitznonmonotonic1991} which extends first-order logic with two modal
operators, $\bfK$ and  $\Not$.
The logic was later extended by Motik and Rosati \shortcite{motikreconciling2010} to
form hybrid MKNF knowledge bases, which support reasoning with ontologies.
We use Knorr et al.'s \shortcite{knorrlocal2011} \mbox{3-valued} semantics for hybrid MKNF
knowledge bases which reason with three truth values: $\ff$ (false), $\uu$ (undefined), and $\tt$ (true) with the
ordering $\ff < \uu < \tt$.
When applied to sets of these truth values, the $min$ and $max$ functions respect this ordering.
A (\mbox{3-valued}) \textit{MKNF structure} is a triple $(I, \M, \N)$ where $I$ is
a (two-valued first-order) interpretation and $\M = \langle M, M_1 \rangle$ and
$\N = \langle N, N_1 \rangle$ are pairs of sets of first-order interpretations
such that $M \supseteq M_1$ and $N \supseteq N_1$.

Hybrid MKNF knowledge bases rely on the standard name assumption under which
every first-order interpretation in an
MKNF interpretation is required to be a Herbrand interpretation with a
countably infinite number of additional constants \cite{motikreconciling2010}.
We use $\Delta$ to denote the set of all these constants.
We use $\phi[\alpha/x]$ to denote the formula obtained by replacing all free
occurrences of variable x in $\phi$ with the term $\alpha$.
Using $\phi$ and $\sigma$ to denote MKNF formulas, Figure \ref{MKNF-evaluate} shows the evaluation of an MKNF
structure.

\begin{figure}
\begin{align*}
    &(I, \M, \N)({p(t_1, \dots ,~t_n)}) \define
    \left\{\begin{array}{ll}
               \tt & \textrm{iff } p(t_1, \dots,~t_n) \textrm{ is true in } I
               \\
               \ff & \textrm{iff } p(t_1, \dots,~t_n) \textrm{ is false in } I
               \\
           \end{array}\right.
    \\
    &(I, \M, \N)({\neg \phi})           \define
    \left\{\begin{array}{ll}
               \tt & \textrm{iff } (I, \M, \N)({\phi}) = \ff \\
               \uu & \textrm{iff } (I, \M, \N)({\phi}) = \uu \\
               \ff & \textrm{iff } (I, \M, \N)({\phi}) = \tt \\
           \end{array}\right.
    \\
    &(I, \M, \N)({\exists x, \phi})     \define max\{(I, \M,
    \N)({\phi[\alpha/x]}) ~|~\alpha \in \Delta\}
    \\
    &(I, \M, \N)({\forall x, \phi})     \define min\{(I, \M,
    \N)({\phi[\alpha/x]}) ~|~\alpha \in \Delta\}
    \\
    &(I, \M, \N)({\phi \land \sigma})   \define min((I, \M, \N)({\phi}), (I,
    \M, \N)({\sigma}))
    \\
    &(I, \M, \N)({\phi \lor \sigma})    \define max((I, \M, \N)({\phi}), (I,
    \M, \N)({\sigma}))
    \\
    &(I, \M, \N)({\phi \subset \sigma}) \define \left\{   \begin{array}{ll}
        \tt & \textit{iff }\vspace{0.3em} (I, \M,
        \N)({\phi}) \geq  (I, \M, \N)({\sigma}) \\
        $\ff$ & otherwise
    \end{array}\right.\\
    &(I, \M, \N)({\bfK \phi})           \define
    \left\{\begin{array}{ll}
               \tt & \textrm{iff  } (J, \MM, \N)({\phi}) = \tt \textrm{ for all
               } J
               \in M
               \\
               \ff & \textrm{iff  } (J, \MM, \N)({\phi}) = \ff \textrm{ for
                   some } J
               \in M_1
               \\
               \uu & \textrm{otherwise}
           \end{array}\right.
    \\
    &(I, \M, \N)({\Not \phi})           \define
    \left\{\begin{array}{ll}
               \tt & \textrm{iff  } (J, \M, \NN)({\phi}) = \ff \textrm{ for
                   some } J
               \in N_1
               \\
               \ff & \textrm{iff  } (J, \M, \NN)({\phi}) = \tt \textrm{ for all
               } J
               \in N
               \\
               \uu & \textrm{otherwise}
           \end{array}\right.
\end{align*}
\caption{the evaluation of an MKNF structure}\label{MKNF-evaluate}
\end{figure}

Intuitively, this logic leverages two sets of interpretations, one for true
knowledge and the other for possibly-true knowledge.
A \Katom{} $\bfK a$ is true if $a$ is true in every ``true'' interpretation,
$\Not a$ holds if $a$ is false in some ``possibly-true'' interpretation.
$\bfK a$ and $\Not a$ are both undefined otherwise.
When we evaluate formulas in this logic, we use a pair of these sets so that
$\Not$-atoms may be evaluated independently from $\bfK$-atoms.
Note that first-order atoms are evaluated under two-valued interpretations,
this is deliberate as, without modal operators, the semantics is essentially
the same as first-order logic.
Also note that a logical implication $\phi \subset \sigma$
may not be equivalent to $\phi \vee \neg \sigma$ if both $\phi$
and $\sigma$ are not first-order formulas.

Knorr et al.\ define their \mbox{3-valued} semantics for the entire language of
MKNF~\shortcite{knorrlocal2011} which subsumes normal hybrid MKNF knowledge bases.
A normal hybrid MKNF knowledge base contains a program and ontology and both are
restricted MKNF formulas which we will now define.

An (MKNF) program $\P$ is a set of (MKNF) rules. A rule $r$ is written in the form $\bfK h \leftarrow \bfK p_0,\dots,~\bfK p_j,~ \Not n_0,\dots,~\Not n_k$ where $h, p_0, n_0, \dots, p_j, n_k $ are function-free first-order atoms
of the form $p(t_0, \dots,~t_n )$ where $p$ is a predicate and $t_0,
    \dots,~t_n$ are either constants or variables.
We call an MKNF formula $\phi$ \textit{ground} if it does not contain variables.
The corresponding MKNF formula $\pi(r)$ for a rule $r$ is as follows:
\begin{align*}
    \pi(r) \define \forall \vec{x},~ \bfK h \subset \bfK p_0 \land \dots \land
    \bfK p_j \land \Not n_0 \land \dots \land \Not n_k
\end{align*}
where $\vec{x}$ is a vector of all variables appearing in the rule.
We will use the following abbreviations:
\begin{align*}
    &\pi(\P) \define \bigwedge\limits_{r \in \P} \pi(r)&\\
    &\head(r)         = \bfK h                                
    &\bfK(\bodyn(r))  = \{ \bfK a ~|~ \Not a \in \bodyn(r) \}\\
    &\bodyn(r)        = \{ \Not n_0, \dots,~ \Not n_k \}      
    &\bodyp(r)        = \{ \bfK p_0, \dots,~ \bfK p_j \}       
\end{align*}
    $\pi(\P) \define \bigwedge\limits_{r \in \P} \pi(r)$, 
    $\head(r)         = \bfK h                              $,
    $\bfK(\bodyn(r))  = \{ \bfK a ~|~ \Not a \in \bodyn(r) \}$,
    $\bodyn(r)        = \{ \Not n_0, \dots,~ \Not n_k \}    $,
    $\bodyp(r)        = \{ \bfK p_0, \dots,~ \bfK p_j \}     $.

A {\em (normal) hybrid MKNF knowledge base} (or knowledge base for short) $\KB
    = (\OO, \P)$ consists of an ontology $\OO$, which is a decidable
description
logic (DL) knowledge base translatable to first-order logic, and a program
$\P$.
We use $\pi(\OO)$ to denote the translation of $\OO$ to first-order logic and
write $\pi(\KB)$ to mean $\pi(\P) \land \bfK \pi(\OO)$.
A {\em (\mbox{3-valued}) MKNF interpretation (pair)} $(M, N)$ is a pair of sets of
first-order interpretations where $\emptyset \subset N \subseteq M$.
We say an MKNF interpretation $(M, N)$ \textit{satisfies} a knowledge base $\KB
    = (\OO, \P)$ if for each $I \in M$,
$(I, \MN, \MN) (\pi (\KB)) = \tt$.

\begin{definition} \label{model}
    A \mbox{3-valued} MKNF interpretation pair $(M, N)$ is a
    \textit{(\mbox{3-valued}) MKNF model} of a normal hybrid MKNF knowledge base
    $\KB$
    if $(M, N)$ satisfies $\pi(\KB)$  and for every \mbox{3-valued} MKNF
    interpretation pair $(M', N')$ where $M \subseteq M'$, $N \subseteq N'$,
    $(M, N) \not= (M', N')$, and we have some $I \in M'$ s.t. $(I, \langle M', N'
        \rangle, \MN)(\pi(\KB)) \not= \tt$.
\end{definition}

Note that the second condition of our definition differs slightly from the
original definition from Knorr et al.\ \shortcite{knorrlocal2011}. They require that
$M' = N'$ if $M = N$; While Knorr et al.'s definition applies to all MKNF
formulas, this condition is not needed when we restrict ourselves to normal
hybrid MKNF knowledge bases~\cite{killen2022}.

Throughout the rest of this paper, we assume that any given hybrid MKNF
knowledge base $\KB = (\OO, \P)$ is {\em DL-safe},
which ensures the decidability by requiring each variable in a rule $r \in \P$
to appear inside some predicate of $\bodyp(r)$ that does not appear in $\OO$.
Throughout this work, and without loss of generality \cite{knorrlocal2011},
we assume rules in $\P$ are ground.

Given a set $S$ of \Katoms{}, we use $\KAK$ and $\OBO{S}$ to denote the following:
\begin{align*}
    \KAK &\define \{ \bfK a ~|~ r \in \P,~ \bfK a \in \head(r) \union 
    \bodyp(r) \union \KB(\bodyn(r)) \}\\
    \OBO{S} &\define \big\{  \pi(\OO) \big\} \union \big\{ a ~|~ \bfK a \in S
    \}
\end{align*}

Knorr et al.\ \shortcite{knorrlocal2011} define when an MKNF interpretation pair $(M,
    N)$ induces an approximation $(T, P) \in \wp(\KAK)^2$ where $T \subseteq P$ if
for
each $\bfK a \in \KAK$:
\begin{itemize}
    \item $\bfK a \in T$ if $\forall I \in M, \mmeval{\bfK a} = \tt$,
    \item $\bfK a \not\in P$ if $\forall I \in M, \mmeval{\bfK a} = \ff$, and
    \item $\bfK a \in P \setminus T$ if $\forall I \in M, \mmeval{\bfK a} =
              \uu$
\end{itemize}

While every MKNF interpretation $(M, N)$ induces a unique approximation $(T, P) \in \wp(\KAK)^2$, in
general, an MKNF interpretation that induces a given approximation is not
guaranteed to exist.
We say an approximation $(T, P) \in \wp(\KAK)^2$ can be \textit{extended} to an MKNF interpretation if there
exists an MKNF interpretation that induces it.
Note that may sometimes refer to an approximation as a model, by this we mean a unique model that the approximation can be extended to.

\section{A Recurrent Approximator}\label{section-main}

The $\Psi$ approximator defined by Liu and You~\shortcite{liuyou2021} blocks the derivation of atoms that appear in the body of a rule with a false head.
However, this blocking only works when such rules are positive.
In \Example{example-running}, if we remove the negative body of rule 1.\, the $\Psi$ approximator computes the well-founded model. 
Using our newly formulated recurrent approximators, we lift Liu and You's operator to
the domain of a tetralattice so that rule derivation can work on all rules,
not just positive ones.

Additionally, we extend the amount of information that can be derived from the
ontology by enabling reasoning with false atoms and the ontology.
In the following, we use $\OBO{P,B}$ as shorthand for $\OBO{P} \union \{ \neg b ~|~ \bfK b \in B \}$ and $\bot$ as shorthand for $a \land \neg a$.

In Figure \ref{fig-operator}, we define a recurrent approximator for hybrid MKNF knowledge bases using
the powerset tetralattice ${\langle \wp(\KAK)^4, \preceq_p^4 \rangle}$.
This operator improves upon prior operators in various ways not discussed here, we direct the reader to our granular comparison~\cite{workshop} for more details.
\begin{figure}
\begin{align*}
    &\Phi(T, F, U, P) \define \Bigg(    \bigcup\limits_{k=0}^2 add_{\,k}(T, P),~
                                       {\KAK} \setminus P,~~~ {\KAK} \setminus T,\\
                                       &\hspace{2.75cm}\bigcup\limits_{k=0}^2 add_{\,k}(P, T) \setminus \bigg(\bigcup\limits_{k=0}^2 extract_{\,k}(T, F, U, P)\bigg)  \Bigg) \\
    &add_{\,0}(T, \argblank)            \define \{ \bfK a \in \KAK ~|~ \OBO{T}
    \models
    a\}
    \\
    &add_{\,1}(T, P)                    \define \{ \bfK a \in \KAK ~|~
    r \in \P,~ \bfK a =
         \head(r),\\ &~~~~~~~~~~~~~~~~~~ \bodyp(r) \subseteq T,~ \bodyn(r) \intersect P = \emptyset
         \}
    \\
    &extract_{\,0}(T, F, \argblank, P)  \define \{ \bfK a \in \KAK ~|~
    	     B
         \subseteq F, \\ & ~~~~~~~~~ B \in filter(F), \OBO{P, B} \not\models \bot, \OBO{T, B}
         \models
         \neg a \}
    \\
    &extract_{\,1}(T, F, \argblank, P)  \define \{ \bfK a \in \KAK ~|~
    	     r \in
         \P, \\ &~~~~ \head(r) \in F, \bodyn(r) \subseteq F, (\bodyp(r) \setminus \{ \bfK
         a
         \})
         \subseteq T \}
\end{align*}
\caption{A recurrent approximator for hybrid MKNF knowledge bases}\label{fig-operator}
\end{figure}
As a reminder, the complement operation leveraged by powerset tetralattices is
the set complement, i.e.,
    $\forall S \in \wp(\KAK),~ \lcomp{S} = \KAK \setminus S$.
For convenience, and without confusion, we write the complement as
$\KAK \setminus S$ from this point forward.
The function $filter(F)$ can be any function $filter(F): \wp(\KAK) \rightarrow
    \wp(\wp(\KAK))$ so long that the following is satisfied.
\begin{align*}
    \forall F, F' \in \wp(\KAK), (F \subseteq F') \implies (filter(F) \subseteq
    filter(F'))
\end{align*}
It is also desirable for $filter(F)$ to be polynomial-time computable and for
its range to be restricted to elements of polynomial size w.r.t.\ some
syntactic measure of $\KB$.
Using the powerset function for filter will result in the most powerful approximator, however, it will not be polynomial.
Less powerful functions trade inference power for tractability while maintaining correctness.

We show that $\Phi$ is a recurrent approximator.
\begin{propositionE}[]\label{monotone}
    For a complete lattice $\langle \LL^4, \preceq^4_{p} \rangle$, the operator
    $\Phi(T, F, U, P)$ is $\preceq^4_p$-monotone.
\end{propositionE}
\begin{proofE}
    Let $(T, F, U, P), (T', F', U', P') \in L^4$ such that $(T, F, U, P)
        \preceq^4_p (T', F', U', P')$.
    By \Proposition{relaxed-monotonicity}, it is sufficient to show $\Phi{(T,
            F, U, P)}_{1, 4} \preceq_p^2 \Phi{(T', F', U', P')}_{1, 4}$.
    It's sufficient to split our task into showing the following
    \begin{align*}
         & (i)~ add_0(T, P) \subseteq add_0(T', P')\\
         &                                                                
        (ii)~ add_1(T, P) \subseteq add_1(T', P')                           \\
         & (iii)~ add_0(P', T') \subseteq add_0(P, T)\\
         &                                                                
        (iv)~ add_1(P', T') \subseteq add_1(P, T)                           \\
         & (v)~ extract_0(T', F', U', P') \subseteq extract_0(T, F, U, P)
                                                                         \\ &
        (vi)~ extract_1(T', F', U', P') \subseteq extract_1(T, F, U, P)
    \end{align*}
    (i and iii) Follow directly from the monotonicity of $\OO$. (ii and iv) We
    show (ii).
    Let $\bfK a \in add_1(T, P)$.
    There exists a rule $r \in \P$ such that $\bodyp(r) \subseteq T$ and
    $\bodyn(r) \intersect P = \emptyset$.
    From $(T, P) \preceq^2_p (T', P')$, we have $\bodyp(r) \subseteq T'$ and
    $\bodyn(r) \intersect P' = \emptyset$, thus,
    $\bfK a \in add_1(T', P')$. The case of (iv) is similar, however, the
    arguments and the relation are flipped.
    That is, we start with $\bodyp(r) \subseteq P'$ and $\bodyn(r) \intersect
        T' = \emptyset$ to conclude
    $\bodyp(r) \subseteq P$ and $\bodyn(r) \intersect T = \emptyset$.

    (v) With $F \subseteq F'$ and because $filter$ is $\subseteq$-monotone, we
    have ${filter(F) \subseteq filter(F')}$.
    Let $B \in filter(F)$ s.t. $B \subseteq F$.
    With $P' \subseteq P$, if we have $\OBO{P,~ B} \not\models \bot$, then we
    have $\OBO{P',~ B} \not\models \bot$.
    With $T \subseteq T'$ and with the monotonicity of $\OO$, we have
    $\OBO{T',~ B} \models \neg a$ when $\OBO{T,~ B} \models \neg a$.

    (vi) Let $\bfK a \in \KAK$ and $r \in \P$ such that $\head(r) \in F$,
    $\bodyn(r) \subseteq F$, and $(\bodyp(r) \setminus \{ \bfK a \}) \subseteq
        T$.
    From $(T, F) \preceq^2_t (T', F')$, we get $\head(r) \in F'$, $\bodyn(r)
        \subseteq F'$, and $(\bodyp(r) \setminus \{ \bfK a \}) \subseteq T'$.
\end{proofE}
The $\Phi$ approximator's stable fixpoints capture the 3-semantics of hybrid MKNF knowledge bases when coupled with the consistency condition used by Liu and You~\cite{liuyou2021} lifted to the tetralattice.
\begin{theoremE}[]\label{main}
    Let $\KB$ be a hybrid MKNF knowledge base and $(T, P) \in \wp(\KAK)^2$ and have the following
    \begin{align*}
        (M, N) &\define (\{ I ~|~ I \models \OBO{T} \}, \{ I ~|~ I \models \OBO{P} \}) \\
        (P^*,U^*) &\define \lfp_{\preceq^2_t}{\Phi({(\cdot, \cdot), ((\KAK \setminus P), T))}}_{1,2}
    \end{align*}
     $(M, N)$ is a \mbox{3-valued} MKNF model of $\KB$ iff
    \begin{align*}
         &i.~ \textrm{$T \subseteq P$},~\\
         &ii.~ \textrm{$(T, (\KAK \setminus P), (\KAK \setminus T), P)$}\\
        &iii.~\textrm{is a fixpoint of $S(\Phi)$ and $\OBO{P^*}$ is consistent.}
    \end{align*}
\end{theoremE}

\begin{proofE}
    ($\Rightarrow$)
    (i) As an MKNF model, we have $N \subseteq M$, thus $T \subseteq P$.
    (ii)
    Because $\Phi{((T, F), (\cdot, \cdot))}_{3}$ is a constant function, we
    need only show
    ${(\lfp_{\preceq^2_t} \Phi{((T, F), (U, \cdot))}_{4}) = P}$.
    We can see that $extract{(T, F, U, \cdot)}_4$ is $\subseteq$-antitone
    \footnote{For more details, see the proof of \Proposition{monotone}}
    , thus to conclude that for all $X \subseteq \KAK$, $extract(T, F, U, X)
        \intersect P = \emptyset$, it is sufficient to show $extract(T, F, U,
        \emptyset) \intersect P = \emptyset$.
    Suppose for the sake of contradiction, we have $\bfK a \in (extract(T, F,
        U, X) \intersect P)$.
    Then either (a) $\OBO{T, B} \models \neg a$ for some $B \subseteq (\KAK
        \setminus P)$ or
    (b) there exists a rule whose head evaluates to false while the body
    evaluates as a non-false value.
    Clearly (b) contradicts the assumption $(M, N)$ is an MKNF model of $\KB$.
    We study (a) more closely.
    $\OBO{P, B}$ is consistent, therefore there exists an interpretation $I \in
        N$ such that $I$ assigns all atoms in $B$ to be false and all atoms in
    $P$ to
    be true.
    With $T \subseteq P$, we have $I \models \OBO{T, B}$, thus $I \models \neg
        a$.
    However, with $I \in N$, we have $I \models a$, a contradiction.
    (iii) $P = P^*$, so $\OBO{P^*}$ is consistent by the initial assumption.
    ($\Leftarrow$)
    To show that $(M, N)$ is an MKNF model of $\KB$, we must show the
    following:
    \begin{enumerate}[(a)]
        \item $(M, N) \mknfmodels \pi(\KB)$
        \item $\forall (M', N')$ s.t. $M \subseteq M'$, $N \subseteq N'$, and
              $(M, N) \not= (M', N')$\\
              $\exists I \in M', \langle I, (M', N'), (M, N) \rangle
                  \not\models
                  \pi(\KB)$
    \end{enumerate}
    (a) Let $(P^*, U^*) = \lfp_{\preceq^2_t}{\Phi({(\cdot, \cdot), ((\KAK
                    \setminus P), T))}}_{1,2}$
    We have $P \subseteq P^*$, thus $\OBO{P}$ is consistent. It follows that
    $(M, N) \mknfmodels \pi(\OO)$.
    Suppose for the sake of contradiction, $(M, N) \not\mknfmodels \pi(\P)$.
    Then there exists a rule $r \in \P$ such that $\bodyp(r) \subseteq P$,
    $\body(r) \intersect T = \emptyset$ and $\head(r) \not\in P$.
    We have $\head(r) \in P^*$, thus $\head(r)$ was remove from $P$ using an
    $extract$ function, a function which only prevents the derivation of atoms
    that
    will result in an inconsistency. Because the set $P^*$ extends $P$ with the
    atoms not blocked by $extract$, we have that $\OBO{P^*}$ is inconsistent, a
    contradiction.

    (b)
    Assume $(M, N) \mknfmodels \pi(\KB)$.
    If $extract$ were to remove atoms computed, the above would not hold, thus we
    have $P = P^*$.
    It follows that
    \begin{align*}
        (T, P) = \Big( & \lfp_{\preceq^2_t} \Phi((\cdot, \cdot), (\KAK
        \setminus
        \boxed{T}), \boxed{P})_{1,2},
        \\
                       & \lfp_{\preceq^2_t} \Phi((\cdot, \cdot), (\KAK
        \setminus \boxed{P}),
        \boxed{T})_{1,2}\Big)_{1,4}
    \end{align*}
    That is if we remove the $extract$ function from the operator, the result
    is identical.
    The remainder of the proof is very similar to the $\Psi$ approximator that $\Phi$ embeds ~\cite{liuyou2021}.
\end{proofE}

There are more stable fixpoints of $S(\Phi)$ than the three-valued models of a given knowledge base.
Because model generation is NP-hard~\cite{liuthreevalued2017}, a polynomial approximator cannot precisely capture these models as stable fixpoints.
Instead, we use a consistency condition to check whether a stable fixpoint is an intended model.

In the following, we revisit the example from \Section{section-introduction}.
\begin{example}
    Let $\KB = (\OO, \P)$ where $\OO = \neg c$ and $\P = $
    \begin{align*}
        & 1.\hspace{1.5em} \bfK a\hspace{0.25em} \leftarrow \Not a'.
         && 2.\hspace{1.5em} \bfK a' \leftarrow \Not a.
       \\
        & 3.\hspace{1.5em} \bfK c\hspace{0.25em} \leftarrow \bfK a,~ \Not b.
   \end{align*}
   When we apply the stable operator $S(\Phi)$ to compute the least fixed point we get the following sequence.
   (For brevity, we omit the third parameter $U$ because it does not play a role in our approximator)
   \begin{enumerate}[i.]
    \item $S(\Phi)(\emptyset, \emptyset, \argblank, \KAK) = (\emptyset, \emptyset, \argblank, \{ a, a' \})$ \\ The atoms $a$ and $a'$ are possibly true, the derivation of $c$ is blocked by $extract_0$.
    \item $S(\Phi)(\emptyset, \emptyset, \argblank, \{ a, a' \}) = (\emptyset, \{ c \}, \argblank, \{ a, a' \})$ \\ the inner, recurrent part of the approximator inverts $P$ from the previous iteration to establish that $c$ is false.
    \item $S(\Phi)(\emptyset, \{ c \}, \argblank, \{ a, a' \}) = (\emptyset, \{ c \}, \argblank, \{ a' \})$\\ Now that $c$ is false, $extract_1$ will block the derivation of $a$
    \item $S(\Phi)(\emptyset, \{ c \}, \argblank, \{ a' \}) = (\{ a' \}, \{ c, a \}, \argblank, \{ a' \})$ \\ Finally, $a$ is established as false and $a'$ is derived as true.
   \end{enumerate}
   Now, let's append the rule $\bfK b \leftarrow \bfK a$ to $\P$
   There is no longer a well-founded model of $\KB$, and $S(\Phi)$ has three stable fixpoints.
   \begin{align*}
     &i.\hspace{0.37cm} S(\Phi)(\emptyset, \{ c \}, \KAK, \{ a, a', b \}) &
     ii.~ S(\Phi)(\{ a, b \}, \{ c, a' \}, \{ c, a' \}, \{ a, b \})\\
     &iii.~ S(\Phi)(\{ a' \}, \{ c, a, b \}, \{ c, a, b \}, \{ a' \})
   \end{align*}
   The fixpoints ii.\ and iii.\ both correspond to three-valued models of $\KB$, however, i.\ is not an MKNF model because rule 1.\ is not satisfied. Note that ii.\ and iii.\ both satisfy the consistency condition in \Theorem{main}, whereas i.\ does not. It is through this property check that we identify which stable fixpoints correspond to intended models 
\end{example}

The method we use to embed false information in approximations has more advantages than the ability to block the derivation of atoms that appear in rules with a false head.
The $\Psi$ approximator is limiting to considering an atom $a$ to be false if $\OBO{T} \models \neg a$.
Our approximator $\Phi$ remembers which atoms are false and therefore it can also interleave the ontology and the program when determining whether an atom is false.
\begin{example}
    Let $\KB = (\OO, \P)$ where $\OO = (x \lor y) \land (\neg c \iff (\neg x \lor \neg y))$ and $\P = $
    \begin{align*}
        & 1.\hspace{1.5em} \bfK c\hspace{0.25em} \leftarrow \bfK b,~ \Not a.\\
        & 2.\hspace{1.5em} \bfK c\hspace{0.25em} \leftarrow \Not c'.
        && 3.\hspace{1.5em} \bfK c' \leftarrow \Not c.\\
        & 4.\hspace{1.5em} \bfK b\hspace{0.25em} \leftarrow \Not b'.
        && 5.\hspace{1.5em} \bfK b' \leftarrow \Not b.\\
        & 6.\hspace{1.5em} \bfK x \leftarrow \bfK x.
        && 7.\hspace{1.5em} \bfK y \leftarrow \bfK y.
    \end{align*}
    Let's compute the least fixed point of the stable operator $S(\Phi)$.
    We assume that $filter(F) = \wp(F)$.
    After one iteration, we conclude that both $x$ and $y$ are false.
    On subsequent iterations, $x$ and $y$ appear in the parameter $F$.
    However, $c$ is not false as it can be derived through rules $2$ and $3$.
    In $extract_0(T, F, U, P)$, we cannot use the set $B = \{a, b\}$ because $\OBO{B}$ is inconsistent.
    However, it is safe to use the singleton set $B = \{ a \}$ to obtain $\OBO{P,B} \models \neg c$.
    Critically, this inference will hold before $c \in P$, thus the second iteration of stable revision will establish $c$ as false.
    If $filter(F) = \emptyset$, then this inference would not occur.
    Once $c$ is established as false, we can infer that $a$ must also be false to satisfy rule 1.
    The least stable fixpoint is $(\{ c', b' \}, \{ c', b' \})$. This approximation also corresponds to the well-founded model. 
\end{example}

While stable revision captures 3-valued hybrid MKNF semantics, we can also use it to refine arbitrary approximations so that they are closer to models w.r.t.\ the number of atoms that need to their truth value changed.

\begin{corollaryE}[]
    Let $(T, P) \in \wp(\KAK)^2$ and let $(M, N)$ be a \mbox{3-valued} MKNF model of a
    hybrid MKNF knowledge base $\KB$ that induces $(T^*, P^*)$ and suppose we have $(T,
        P)
        \preceq^2_p (T^*, P^*)$  (resp. $(T^*, P^*) \preceq^2_p (T, P)$).
    The following holds
    \begin{align*}
        {S(\Phi)({(T, P)}^4)}_{1, 4}  \preceq^2_p (T^*, P^*) &&
        (\textrm{resp. }(T^*, P^*)    \preceq^2_p {S(\Phi)({(T, P)}^4)}_{1,
        4})
    \end{align*}
\end{corollaryE}
\begin{proofE}
    With \Proposition{S-monotone} and \Proposition{monotone}, $S(\Phi)$ is
    $\preceq^4_p$-monotone.
    Thus
    \begin{align*}
        S(\Phi)({S(\Phi)({(T, P)^4})})          & \preceq^4_p S(\Phi)({(T^*,
        P^*)}^4)                                                             \\
        (\textrm{resp. }S(\Phi)({(T^*, P^*)}^4) & \preceq^4_p
        S(\Phi)({S(\Phi)({(T, P)^4})})
    \end{align*}
    By \Theorem{main}, $S(\Phi)({(T^*, P^*)}^4) = {(T^*, P^*)}^4$. Finally,
    $\preceq^4_p$ embeds the $\preceq^2_p$ relation (See
    \Definition{orderings}).
\end{proofE}


The well-founded operators for hybrid MKNF knowledge bases~\cite{Ji17,killenunfounded} can easily be embedded within a solver.
This is partly because the operators are monotone increasing.
Stable revision is not so easy to integrate into a solver, in general, a fixpoint may not exist for an infinite lattice.
However, for a monotone increasing (or decreasing) operator, we're guaranteed
to reach a fixpoint when we repeatedly apply an operator to an arbitrary
lattice element; this property is needed by solvers which handle arbitrary approximations.

Another use for recurrent approximators is a method to turn any approximator into an increasing approximator.
Given a recurrent approximator $o$ over a powerset tetralattice $\langle
    \wp(\LL)^4, \preceq_p^4 \rangle$, we can easily define
$\preceq^4_p$-monotone
increasing and decreasing variants of $o$ which we denote as $o^{+}$ and
$o^{-}$ respectively.
\begin{align*}
    &o^{+}(T, F, U, P)_{1,4}  \define\\
    &~~~~~~~~~\Big(
    (o(T, F, U, P)_{1} \union \lcomp{U}),
    (o(T, F, U, P)_{4} \setminus F)\Big)    \\
    &o^{-}(T, F, U, P)_{1,4}  \define \\
    &~~~~~~~~~ \Big(
    (o(T, F, U, P)_{1} \intersect U),
    (o(T, F, U, P)_{4} \union \lcomp{F})\Big)
\end{align*}
Both $o^{+}$ and $o^{-}$ are recurrent approximators.
It's noteworthy that their increasing/decreasing properties also carry over to
their stable revision operators.
\begin{remark}
    For a recurrent approximator $o(T, F, U, P): \LL^4 \rightarrow \LL^4$ over a
    powerset tetralattice $\langle \wp(\LL)^4, \preceq_p^4 \rangle$, the
    operators
    $S(o^+)$ and $S(o^-)$ are $\preceq^4_p$-monotone increasing and decreasing
    respectively.
\end{remark}
It may be possible to formulate similar $o^{+}$ and $o^{-}$
approximators for any tetralattice (not just powerset tetralattices), however,
we do not explore that here.

The increasing variant of a recurrent approximator can be interleaved with any other propagation method in a solver without the worry that a fixpoint will not be reached.

\section{Summary}\label{section-discussion}

We introduced recurrent approximators, operators defined on a tetralattice that can store false information computed in previous iterations of stable revision.
We demonstrated how these operators can be applied to hybrid MKNF knowledge bases by defining a new approximator that has fewer unintended stable fixpoints than previous approximators.
This approximator can be viewed as a unification of developments for hybrid MKNF knowledge bases that use well-founded operators \cite{Ji17,killenunfounded} and the approximators defined for hybrid MKNF knowledge bases \cite{liuyou2021}.
The new approximator is more precise than both prior works.

We expand the class of hybrid MKNF knowledge bases that have a known polynomial algorithm to compute the well-founded model when one exists.
Our extended AFT widens the scope of applications of AFT. Stable revision in current AFT can capture reasoning with a single system, but if such a system is integrated with other systems, where ``stale information'' comes from other reasoning contexts, our framework is needed.
While in this work we focus on hybrid MKNF, these techniques could be applied with other hybrid reasoning systems.

The problem of how to ground a hybrid MKNF KB has not been addressed in the literature.
The least fixpoint of our proposed approximator can be used as a basis for grounding MKNF rules because it only makes well-founded inferences. 
Our proposed approximator also provides a basis for a more powerful constraint propagator for building a solver - it induces a family of well-founded operators that can be used to replace those from Ji et al.\ \cite{Ji17} while preserving the soundness and completeness of the solver (THM 4.3 from Ji et al.\ \cite{Ji17}).

\section{Future Work}\label{section-future}

One limitation of the approximator defined in \Section{main}, is that it can
only block the derivation of atoms that appear in the body of a rule whose head
is false if the rest of the rule is true.
If there exists a rule whose head is false but there are multiple atoms that
are undefined, then the operator will not be able to compute a well-founded
approximation.
\begin{example}
    Let $\KB = (\OO, \P)$ be a hybrid MKNF knowledge base defined as
    $\OO = \neg a$
    and where $\P$ contains the following rules
    \begin{align*}
        \bfK b\hspace{0.3em} & \leftarrow \Not b'         \\
        \bfK b'              & \leftarrow \Not b          \\
        \bfK c\hspace{0.3em} & \leftarrow \Not c'         \\
        \bfK c'              & \leftarrow \Not c          \\
        \hspace{1px}                                      \\
        \bfK a\hspace{0.3em} & \leftarrow \bfK b,~ \bfK c \\
        \bfK a\hspace{0.3em} & \leftarrow \Not b
    \end{align*}
\end{example}
The example above has two \mbox{3-valued} MKNF models that induce the
approximations $(\{ c' \}, \{  b, b', c' \})$ and $(\{ c', b \}, \{  b, c' \})$.
However, the least stable fixpoint our operator computes is
$(\emptyset, \{ b', b, c', c \})$.
Lookahead can provide an avenue of further refinement for our operator.
If we can quickly test that no MKNF model assigns $\bfK b$ to be false, then we
can block the derivation of $\bfK c$ and arrive at a \mbox{3-valued} MKNF model.

\section{Acknowledgements}

We would like to acknowledge and thank Alberta Innovates and Alberta Advanced
Education for their direct financial support of this research.

\bibliographystyle{plain}
\bibliography{document}

\newpage
\appendix

\section{Appendix: Proofs}\label{appendix}

\printProofs

\section{Removing the Complement Requirement}\label{appendix-no-complement}

We briefly sketch an alternative definition of recurrent approximators that
avoids the need to define a complement operation on the lattice.
Given a lattice $\langle L, \preceq_{\LL} \rangle$, we define the tetralattices
\begin{align*}
    \langle {\LL}^4, \preceq^4_{t*} \rangle, \langle {\LL}^4, \preceq^4_{p*}
    \rangle
\end{align*}
Where $\preceq_{t*}^4$ is defined for any two 4-tuples $(T, F, U, P), (T', F',
    U',
    P') \in {\LL}^4$ such that the following are equivalent
\begin{itemize}
    \item $((T, F), (U, P)) \preceq_{t*}^{4} ((T', F'), (U', P'))$,
    \item $(T, F) \preceq^2_{t} (T', F') \land (U, P) \preceq^2_{t} (U', P')$,
          and
    \item $T \preceq_{\LL} T' \land F \preceq_{\LL} F'	\land U \preceq_{\LL} U' \land P
              \preceq_{\LL} P'$
\end{itemize}
The $\preceq^4_{p*}$ ordering is defined such that the following are equivalent
\begin{itemize}
    \item $((T, F), (U, P)) \preceq_{p*}^{4} ((T', F'), (U', P'))$,
    \item $(T, F) \preceq^2_{p} (T', F') \land (U', P') \preceq^2_{p} (U, P)$,
    \item $(T, P) \preceq^2_{p} (T', P')$, and
    \item $T \preceq_{\LL} T' \land F' \preceq_{\LL} F \land U \preceq_{\LL} U' \land P'
              \preceq_{\LL} P$
\end{itemize}

With these new orderings, a \textit{$*$-recurrent approximator} is a defined as
a $\preceq^4_{p*}$-monotone operator $o$ over the complete lattice $\langle
    {\LL}^4, \preceq^4_{p*} \rangle$ such that
\begin{align*}
    o(T, F, U, P)_{2,3} \define (F, U)
\end{align*}

Unlike our preferred definition of recurrent approximators
(\Definition{orderings}), the tetralattices ${\langle {\LL}^4, \preceq^4_{t*}
    \rangle}$ and ${\langle {\LL}^4, \preceq^4_{p*} \rangle}$ are not bilattices
formed from the bilattice ${\langle {\LL}^4, \preceq^2_{t} \rangle}$.
For this reason, the application of AFT is less immediate.





\end{document}